\documentclass[conference]{IEEEtran}
\IEEEoverridecommandlockouts

\usepackage{cite}
\usepackage{amsmath,amssymb,amsfonts}
\usepackage{algorithmic}
\usepackage{graphicx}
\usepackage{textcomp}
\usepackage{xcolor}
\usepackage{subcaption}
\usepackage{booktabs}
\def\BibTeX{{\rm B\kern-.05em{\sc i\kern-.025em b}\kern-.08em
    T\kern-.1667em\lower.7ex\hbox{E}\kern-.125emX}}
\begin{document}

\title{sensVLA: Spatially-Grounded Vision-Language-Action Model for Autonomous Wheel Loader}



\author{
\IEEEauthorblockN{Gopi Krishna Erabati, Bjarne Johannsen, Angus Stewart, Vardeep Singh Sandhu}
\IEEEauthorblockA{
\textit{sensmore GmbH} \\
Berlin, Germany
}
}

\maketitle

\begin{abstract}
Autonomous wheel-loader control requires joint reasoning over task semantics, egocentric vision, proprioception, and 3D scene geometry. We present sensVLA, a Vision–Language–Action (VLA) architecture that combines a Qwen3-2B Vision–Language Model (VLM) with a fully trainable transformer action expert trained by flow-matching velocity regression. sensVLA routes Bird's-Eye-View (BEV) features, extracted from fused front and rear lidar, directly to the action expert through a dedicated cross-attention pathway, while the VLM consumes front and rear RGB views to provide task-conditioned semantic context. This design decouples spatial grounding from linguistic reasoning while preserving interaction between both streams at decision time. The expert predicts six action dimensions: longitudinal velocity, steering, body-frame displacement, arm rate, and bucket rate. On a real-world dataset from a wheel loader, sensVLA reaches aggregate per-step parity with a strong camera-only baseline and reduces longitudinal-velocity RMSE by $28\%$ and displacement error by $9\%$ on loading-centric scenarios. It also degrades $29\%$ less when the camera stream is corrupted or removed, evidencing that explicit spatial grounding improves accuracy and fault-tolerance for heavy-equipment autonomy.
\end{abstract}

\begin{IEEEkeywords}
Vision-Language-Action Model, BEV Perception, Autonomous Wheel Loader,
Flow Matching, Imitation Learning
\end{IEEEkeywords}

\section{Introduction}
\label{sec:intro}

Autonomous wheel loaders face a demanding control problem: the policy must navigate unstructured terrain, coordinate both drivetrain and hydraulic actuators, and respect scene semantics ranging from rock piles to tunnel walls. Recent Vision-Language-Action (VLA) models~\cite{rt2,openvla,pi0} suggest that large vision-language backbones can provide these semantic priors, provided that the action head remains expressive enough for continuous, high-frequency control.

However, directly transferring autoregressive image-text VLAs to heavy equipment is difficult for two reasons. First, outdoor earthmoving tasks require metric spatial reasoning, such as distance to the pile face, bucket-to-ground clearance, and berm heading, that RGB tokens alone capture poorly. A lidar-derived BEV representation provides this missing geometry. Second, real-world machine data is limited relative to the size of modern VLMs, making full fine-tuning statistically inefficient. Parameter-efficient fine-tuning (PEFT) is therefore essential.

These observations motivate a key architectural question: where should BEV features enter the network? We propose sensVLA, a VLA architecture built around a Qwen3-2B VLM and a fully trainable transformer action expert that consumes front and rear RGB cameras and front and rear lidar. Bidirectional perception is important because a wheel loader routinely reverses between pile and dump, and a front-only view cannot capture hazards behind the machine during hauling. Instead of passing BEV features through the language decoder, sensVLA routes a frozen PointPillars BEV embedding to the action expert through a parallel cross-attention pathway while also using task-level context from the VLM. The expert interleaves self-attention with heterogeneous cross-attention, where each cross-attention layer attends to VLM features, projected BEV features, or their concatenation. This design preserves the spatial resolution of the BEV grid without competing for the VLM token budget and yields robust performance on a real-world loading-and-hauling dataset collected from a production wheel loader. It also improves robustness under sensor degradation, particularly when camera input is degraded or unavailable.

Our main contributions are as follows.

\begin{itemize}
    \item We introduce sensVLA, a VLA architecture that decouples geometric
    grounding from language reasoning by routing BEV lidar features into a
    fully trainable action expert through heterogeneous cross-attention,
    while leveraging a LoRA-adapted VLM for task-level context.

    \item We formulate action generation as a flow-matching velocity
    regression over temporally chunked 6-DoF predictions, enabling
    continuous multi-actuator control without discretisation artifacts.

    \item We present a staged, parameter-efficient training strategy with
    adapter unfreezing, state-history dropout, and EMA validation to
    mitigate overfitting under data-scarce conditions.

    \item We empirically show that BEV grounding improves loading-centric
    accuracy over a strong camera-only baseline and degrades markedly less
    under simulated camera loss, establishing spatial grounding as both an
    accuracy and a fault-tolerance lever for heavy-equipment VLAs.
\end{itemize}

\section{Related Work}
\label{sec:related}

Prior work most relevant to sensVLA falls into two areas: BEV-based perception for metric grounding and vision-language-action models for semantically informed control. In autonomous driving, BEV representations have become a standard way to encode scene geometry in an ego-centric metric frame. PointPillars, Lift-Splat-Shoot, BEVFormer, BEVFusion, and TransFuser show that lidar- and multi-view-based BEV features support strong perception and control-oriented reasoning~\cite{pointpillars,lss,bevformer,bevfusion,transfuser}. Unlike these methods, sensVLA routes BEV features directly to the action generator rather than keeping them inside the perception stack.

In parallel, VLA and generalist robot-policy work has shown that transformer backbones can combine language, vision, and action across tasks~\cite{gato,palme,rt1,rt2,openx,octo,openvla}. Parameter-efficient adaptation methods such as LoRA are especially useful when in-domain robot data is limited~\cite{hu2022lora}, while recent action heads move beyond discrete tokenization toward chunked and continuous prediction~\cite{act,diffusion_policy,pi0}. sensVLA combines these directions by pairing VLM-based semantic reasoning with explicit BEV grounding for end-to-end articulated heavy-machine control, rather than relying on the modular loading pipelines used in prior wheel-loader systems~\cite{lhd_autoload,auto_loading_unknown_material,world_model_loader}.

\section{System Overview and Data Pipeline}
\label{sec:data}

The proprietary dataset consists of lidars, RGB cameras, and IMU sensors. Each sensor is stored asynchronously and during training, temporally synchronized to the control timestamps.

Each sample consists of the RGB images, a short natural-language task prompt, 0.5s of accumulated lidar data, the proprioceptive state, and the action targets. The proprioceptive state consists of forward velocity, steering angle, imu data. The action targets are forward velocity, steering, body-frame displacement, arm rate and bucket rate. Supervision is represented as ten-step action chunks, while conditioning includes the current observation and a ten-step state-history window. Samples with missing packets, invalid synchronization, or incomplete calibration are discarded.

The training corpus consists of two types of scenarios. One where the wheel loader is loading material from a pile and one where the wheel loader is driving through an underground tunnel. To preserve evaluation fidelity, the dataset is split at the recording-sequence level rather than by frame, preventing near-duplicate temporal neighbors from leaking across train
and validation sets.

The result is a compact real-world dataset for articulated loading-and-hauling, with synchronized semantic, geometric, and proprioceptive context in each retained example.




\section{sensVLA Architecture}
\label{sec:method}

\begin{figure}[htbp]
    \centering
    \includegraphics[width=\linewidth]{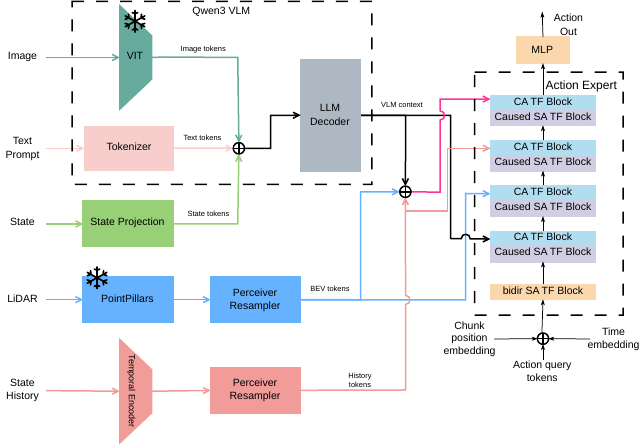}
    \caption{sensVLA Architecture.}
    \label{fig:sensvla_arch}
\end{figure}

 The sensVLA model ingests four modalities: an RGB image, a short text prompt, proprioceptive state including a ten-step history, and a temporally-accumulated lidar point cloud, and produces a chunk of ten future 6-DoF actions as shown in fig.~\ref{fig:sensvla_arch}. The RGB image and text prompt are processed by a LoRA-adapted~\cite{hu2022lora} Qwen3-2B VLM~\cite{Qwen3-VL} to produce task-conditioned contextual representations, while the lidar cloud is encoded into a BEV feature grid by a frozen PointPillars backbone. A fully trainable transformer~\cite{vaswani2017attention} action expert then leverages heterogeneous cross-attention to fuse VLM context and BEV features alongside proprioceptive conditioning, and decodes the resulting representation into temporally chunked actions via flow-matching velocity regression.

\subsection{Image, Text and State Encoding}

We leverage Qwen3-VL-2B-Instruct~\cite{Qwen3-VL} as the VLM backbone. The front and rear RGB frames are each tokenised by the frozen native ViT encoder and the resulting visual tokens are concatenated, while text is tokenised and embedded via the shared embedder. Processing both views through a shared encoder — rather than through separate branches — keeps the pretrained image–text alignment intact and lets the language decoder arbitrate which view to attend to at every layer. We truncate the language decoder to its first 14 transformer blocks: in LLMs, later layers are increasingly specialised toward the pretraining objective and contribute diminishing representational value for downstream visuomotor control, while the early layers retain rich, transferable contextual features. This truncation also keeps the memory footprint tractable and narrows the optimisation gap between the language decoder and the action expert. The current proprioceptive state $s \in \mathbb{R}^{20}$, comprising longitudinal velocity, yaw rate, and 6-channel readings from three IMUs, is projected to the VLM hidden dimension by a small MLP and appended as an additional input token, following~\cite{rt2}. The fused sequence of image, text, and state tokens is processed by the truncated decoder, and the last-layer hidden states serve as the VLM context $C_{\mathrm{vlm}}$ for the action expert.

\subsection{State-History and BEV Encoding}
Beyond the VLM context, the action expert receives two additional conditioning signals that capture temporal dynamics and geometric scene structure respectively.
A ten-step window of past proprioceptive states is encoded by a three-stage convolutional network with progressively increasing channel dimensions and mean-pooled into a single history token $h$. This token is concatenated with $C_{\mathrm{vlm}}$ before entering the cross-attention blocks, providing low-frequency inertial context without inflating the VLM sequence length.
For geometric grounding, we accumulate front- and rear-lidar sweeps over 0.5 s in the front-lidar frame and pass the fused cloud through a pretrained PointPillars ~\cite{pointpillars} encoder. We leverage PointPillars as it achieved comparable performance at much lower runtime which is critical for real time control.  We extract multi-scale features from the encoder's feature pyramid, upsample them to a common spatial resolution with learned convolutions, and concatenate along the channel dimension to obtain a dense BEV feature map $B$. This map is flattened into a sequence of spatial tokens and projected to the expert hidden dimension via a linear layer, with a learned positional embedding added to preserve spatial structure. Crucially, $B$ is routed directly to the action expert through a dedicated cross-attention pathway, decoupling the spatial-to-action mapping from the limited capacity of the LoRA adapters.

\subsection{Action Expert and Flow-Matching Head}
The action expert is a 4-layer, 8-head transformer whose input is the noised action chunk $x_t \in \mathbb{R}^{10 \times 6}$, augmented with a learned positional embedding and a sinusoidal flow-matching time embedding. The first layer is a bidirectional self-attention block that allows all ten query positions to exchange information freely before any autoregressive constraint is imposed. The remaining three layers are causal, each consisting of a causal self-attention followed by a cross-attention block. Rather than attending to the same context in every layer, each cross-attention draws from a different source: the first attends to the VLM context $C_{\mathrm{vlm}}$ for language-visual grounding, the second attends to the projected BEV tokens through a gated residual connection initialised near identity so the spatial pathway amplifies only as training finds it useful, and the third attends to the concatenation of both, letting the softmax distribution select per query which modality is most informative at the fusion stage. This heterogeneous layout exploits the fact that the VLM context is invariant within a forward pass; redirecting a cross-attention layer to BEV features adds spatial grounding at essentially zero additional parameter cost relative to a homogeneous VLM-only expert.

The expert is trained with a flow-matching~\cite{lipman2022flow} velocity regression objective. Given a target action and sampled noise, a linear interpolation parameterised by a scalar $t \sim \mathrm{Beta}(2,5)$ forms the noised input, and the expert learns to predict the ground-truth velocity field under a per-component weighted MSE loss. The loss weights are tuned to balance the heterogeneous physical units across the six action dimensions. At inference, the learned velocity field is integrated from $t{=}1$ to $t{=}0$ using a ten-step midpoint ODE solver, producing continuous multi-actuator commands without discretisation artifacts or VQ-style losses.

\section{Experiments}
\label{sec:experiments}

\subsection{Implementation Details}

We collect data onboard a wheel loader and curate a dataset comprising 200K training chunks and 40K validation chunks. Each sample contains RGB images, LiDAR and IMU measurements. Our policy is built upon Qwen3-2B-VL \cite{Qwen3-VL}, where the language decoder is truncated to 14 transformer blocks. LoRA adapters ($r=16$, $\alpha=16$, dropout $0.05$) are inserted into the final 8 decoder layers. The PointPillars encoder, pretrained on a separate in-house end-to-end task, is kept frozen throughout training. The action expert uses a hidden dimension of 384 and consists of 5 transformer layers (1 bidirectional followed by 4 causal layers) with 8 attention heads and per-block dropout of $0.2$. It attends to 32 BEV tokens through a dedicated cross-attention module. State history spans 10 steps and is compressed into 8 resampled Perceiver tokens. Action prediction is performed over chunks of 10 steps with an action stride of 5, corresponding to a 2 s horizon at 25 Hz. Training uses a batch size of 16 with 2-step gradient accumulation, yielding an effective batch size of 32. We employ AdamW with learning rates of $1\mathrm{e}{-5}$ for the expert and BEV heads, $3\mathrm{e}{-6}$ for the BEV resampler, and $1\mathrm{e}{-6}$ for the LoRA parameters. Optimization follows a cosine decay schedule with 5 \% warm-up and weight decay of $1\mathrm{e}{-3}$. All experiments use bf16 mixed-precision training for 30 epochs. To stabilize optimization, the LoRA adapters remain frozen during the first epoch while the action expert is warmed up against the frozen VLM backbone. Regularization includes BEV token dropout and state-history step/full dropout. As a baseline, we use the same architecture and training setup as sensVLA, but remove the BEV pathway and condition the action expert only on front/rear RGB, text, and proprioceptive inputs.

\subsection{Results and Discussion}

\begin{table}[t]
\centering
\caption{Per-step physical RMSE on the validation set}
\label{tab:e1}
\setlength{\tabcolsep}{4pt}
\renewcommand{\arraystretch}{1.1}
\begin{tabular}{lcccccc}
\toprule
 & \multicolumn{2}{c}{\textbf{Aggregate}}
 & \multicolumn{2}{c}{\textbf{Loading}}
 & \multicolumn{2}{c}{\textbf{Driving}} \\
\cmidrule(lr){2-3}\cmidrule(lr){4-5}\cmidrule(lr){6-7}
Model
 & $dx{+}dy$ & $v_x$
 & $dx{+}dy$ & $v_x$
 & $dx{+}dy$ & $v_x$ \\
 & (m) & (m/s) & (m) & (m/s) & (m) & (m/s) \\
\midrule
Baseline
 & 0.107 & 1.134
 & 0.106 & 0.996
 & 0.107 & 1.281 \\
\textbf{sensVLA (ours)}
 & \textbf{0.101} & \textbf{0.886}
 & \textbf{0.097} & \textbf{0.717}
 & \textbf{0.105} & \textbf{1.054} \\
\bottomrule
\end{tabular}
\end{table}

Aggregated over the 40K validation chunks (Table~\ref{tab:e1}), sensVLA reduces longitudinal-velocity RMSE from $1.134$ to $0.886$ m/s ($-21.9\%$) and displacement-$dx$ RMSE from $9.52$ to $8.98$ cm ($-5.7\%$), while steering and lateral-$dy$ errors remain statistically identical. The gap widens once the validation set is stratified by task group: on the loading group sensVLA reduces $v_x$ RMSE from $0.996$ to $0.717$ m/s ($-28.0\%$) and $dx$ RMSE from $9.27$ to $8.40$ cm ($-9.4\%$); on the driving group $v_x$ also improves, from $1.281$ to $1.054$ m/s ($-17.7\%$), while $dx$ remains effectively flat. The loading split, where metric cues such as pile face geometry and the bucket-approach corridor disambiguate intent that RGB alone cannot capture, is precisely where lidar grounding pays off the most.

Rolling these per-step residuals out over the 2-second prediction horizon preserves the same ordering: on the loading group, final displacement error drops from $0.538$ m to $0.521$ m ($-3.1\%$), confirming that the per-step gains do not wash out under integration.

\begin{table}[t]
\centering
\caption{Robustness under input ablation. ``Agg'' is the mean
per-component RMSE in z-normalised action space; $\Delta\,\times$ is the
ratio to each model's \emph{full} anchor (lower is better).}
\label{tab:e2}
\setlength{\tabcolsep}{5pt}
\renewcommand{\arraystretch}{1.1}
\begin{tabular}{lcccc}
\toprule
 & \multicolumn{2}{c}{\textbf{Baseline}}
 & \multicolumn{2}{c}{\textbf{sensVLA (ours)}} \\
\cmidrule(lr){2-3}\cmidrule(lr){4-5}
Input variant & Agg & $\Delta\,\times$ & Agg & $\Delta\,\times$ \\
\midrule
full (anchor)    & 1.017 & $1.00\times$ & 1.022 & $1.00\times$ \\
img-blur         & 1.06 & $1.04\times$ & 1.022 & $1.00\times$ \\
no-state-hist    & 1.175 & $1.16\times$ & 1.090 & $1.07\times$ \\
no-bev           & ---   & ---          & 1.206 & $1.18\times$ \\
\midrule
\textbf{no-cam}  & 2.378 & $2.34\times$ & \textbf{1.704} & \textbf{$1.67\times$} \\
\bottomrule
\end{tabular}
\end{table}

\textbf{Robustness under sensor degradation}. Table~\ref{tab:e2} reports aggregate normalised RMSE under four input perturbations, relative to the unperturbed anchor. Under mild photometric corruption (\emph{img-blur}) both models are unaffected ($\sim1.00\times$). Under simulated camera loss (\emph{no-cam}), the baseline degrades by $2.34\times$ while sensVLA degrades by only $1.67\times$; the effect is sharpest on $v_x$ RMSE, which explodes from $1.13$ to $9.05$ m/s for the baseline but only from $0.89$ to $2.47$ m/s for sensVLA ($3.7\times$ smaller collapse). The converse sanity check (\emph{no-bev}) lands at $1.18\times$, confirming that BEV is actively used but not the sole load-bearing modality. BEV spatial grounding therefore provides a measurable, asymmetric fallback: the action expert can keep predicting metric-consistent motion when the RGB stream is lost, which a camera-only VLA cannot by construction.

\begin{figure}[t]
\centering
\includegraphics[width=\linewidth]{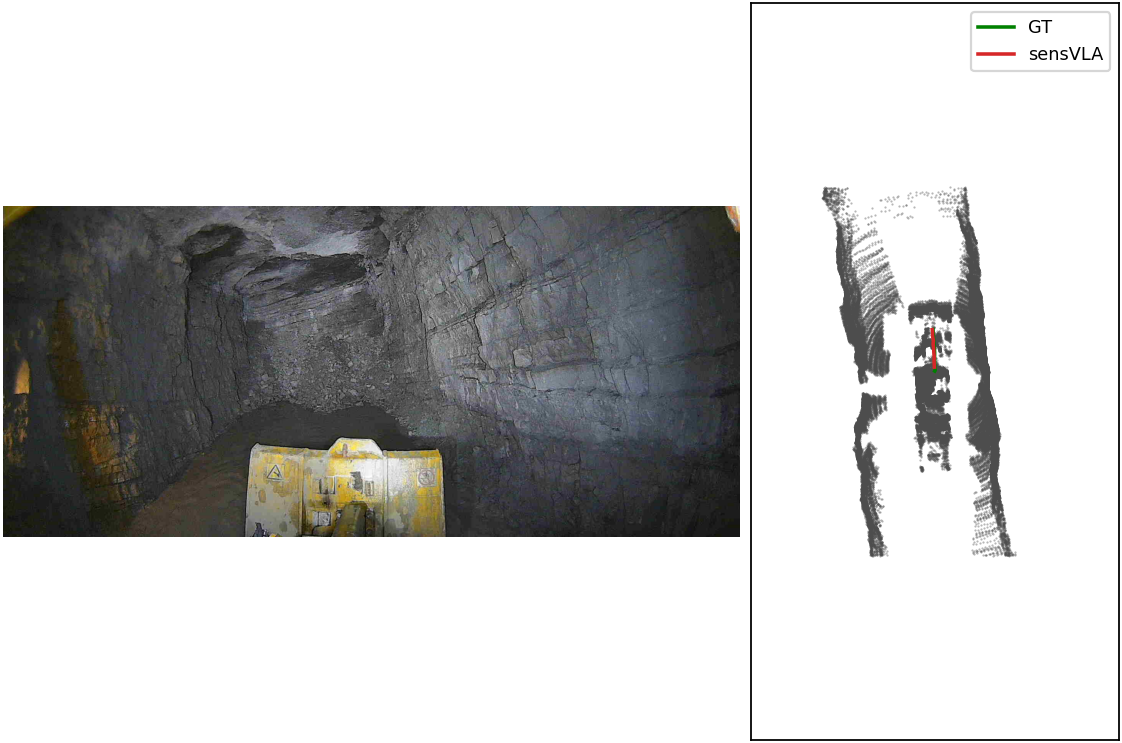}\\[3pt]
\includegraphics[width=\linewidth]{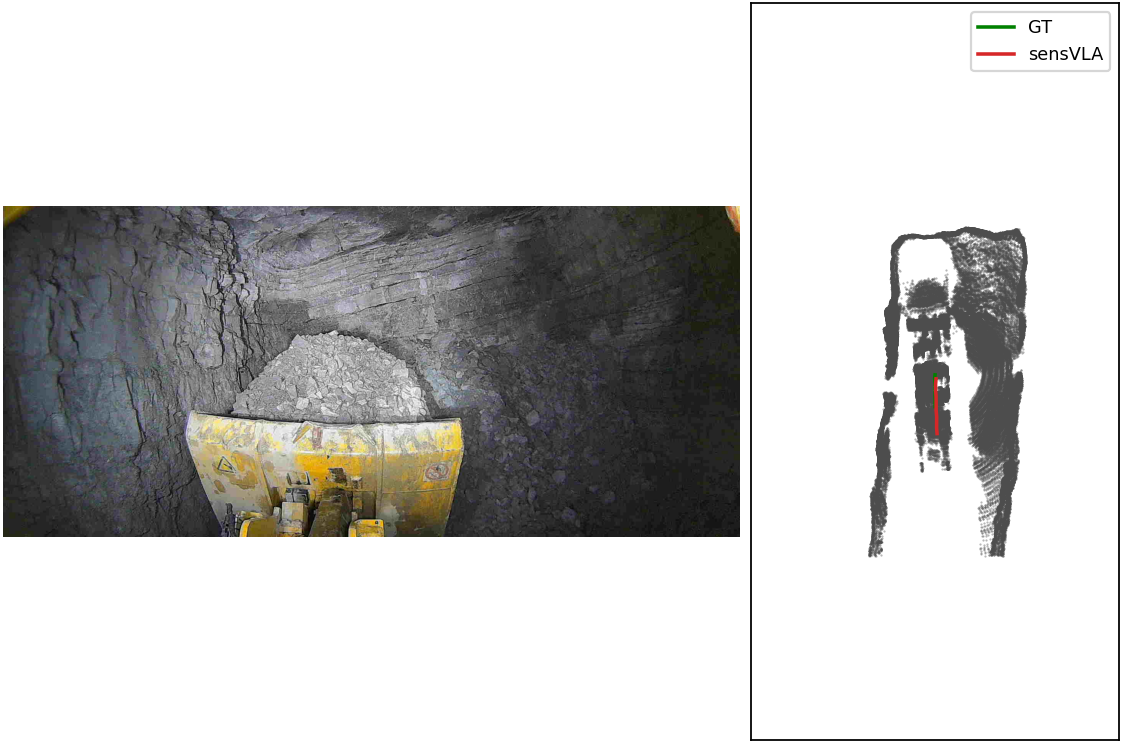}
\caption{Qualitative predictions on two validation frames (loading approach). On the BEV, the ground-truth 2\,s ego-path is drawn in \textcolor{green!60!black}{green} and sensVLA's prediction in \textcolor{red}{red}.}
\label{fig:qualitative}
\end{figure}

Fig.\ref{fig:qualitative} visualises sensVLA's predictions on two validation frames where metric scene geometry is the decisive cue: a forward pile approach (left) and a reverse manoeuvre after a bucket interaction (right). In both cases the red predicted path stays within centimetres of the green ground truth, including through the lateral turn, despite the front camera being partly occluded by material (right) and rear because of no light. The tight red--green agreement mirrors the quantitative gains of Tables\ref{tab:e1}--\ref{tab:e2} and evidences that BEV grounding lets the expert commit to the geometrically correct trajectory rather than averaging across plausible monocular hypotheses.

\section{Conclusion and Future Work}
\label{sec:conclusion}

We presented sensVLA, a vision-language-action model for autonomous wheel loaders that injects front and rear lidar BEV features directly into a flow-matching action expert while leaving a LoRA-adapted Qwen3-VL backbone to specialize in language and RGB understanding. Routing geometry to the expert rather than through the VLM preserves the pretrained language-visual representation and places spatial grounding where trainable capacity is highest. sensVLA matches a strong camera-only baseline in aggregate per-step error and improves substantially on loading-centric slices, including a $28\%$ reduction in $v_x$ RMSE and a $9.4\%$ reduction in $dx$ RMSE on the loading group, while degrading $29\%$ less when the camera stream is ablated. These results support our main claim: explicit spatial grounding fused at the decision stage, rather than inside the language decoder, provides a data-efficient and fault-tolerant recipe for VLA-based heavy-equipment autonomy. Future work includes co-training the PointPillars encoder with the expert, extending temporal BEV fusion over longer histories and leverage sensVLA for multi-tasking.
\bibliographystyle{IEEEtran}
\bibliography{references.bib}

\end{document}